# TeaLeafVision: An Explainable and Robust Deep Learning Framework for Tea Leaf Disease Classification

Rafi Ahamed
*Dept. of CSE*
East West University
Dhaka, Bangladesh
fidaahamed15@gmail.com

Sidratul Moon Nafsin
*Dept. of BBA*
East West University
Dhaka, Bangladesh
nafsinsidratulm@gmail.com

Md. Abir Rahman
*Dept. of CSE*
East West University
Dhaka, Bangladesh
abirrahman.sakin@gmail.com

Tasnia Tarannum Roza
*Dept. of CSE*
East West University
Dhaka, Bangladesh
tasniatarannumroza28@gmail.com

Munaia Jannat Easha
*Dept. of CSE*
East West University
Dhaka, Bangladesh
jannateasha004@gmail.com

Abu Raihan
*Dept. of CSE*
East West University
Dhaka, Bangladesh
abu.raihan2295@gmail.com

*Abstract*— As the world's second most-consumed beverage after water, tea is not just a cultural staple but a global economic force of profound scale and influence. More than a mere drink, it represents a quiet negotiation between nature, culture, and the human desire for a moment of reflection. So, the precise identification and detection of tea leaf disease is crucial. With this goal, we have evaluated several Convolutional Neural Networks (CNN) models, among them three shows noticeable performance including DenseNet201, MobileNetV2, InceptionV3 on the teaLeafBD dataset. teaLeafBD dataset contains seven classes, six disease classes and one healthy class, collected under various field conditions reflecting real-world challenges. Among the CNN models, DenseNet201 has achieved the highest test accuracy of 99%. In order to enhance the model reliability and interpretability, we have implemented Gradient-weighted Class Activation Mapping (Grad-CAM), occlusion sensitivity analysis and adversarial training techniques to increase the noise resistance of the model. Finally, we have developed a prototype in order to leverage the model's capabilities on real life agriculture. This paper illustrates the deep learning model's capabilities to classify the disease in real-life tea leaf disease detection and management.

*Keywords*— *Tea Leaf, Leaf Disease Classification, Grad-CAM, Occlusion sensitivity, Adversarial training, Deep learning, CNN, Explainable AI*

## I. INTRODUCTION

Tea is a popular beverage worldwide and a significant commercial crop in many nations. [1] Tea plantations' quality and productivity are heavily reliant on the health of their leaves, which are highly open to being affected to a variety of fungal, bacterial, and pest-borne diseases. Common diseases like blister blight, grey blight, and red rust cause serious damage to tea leaves, resulting in significant economic losses. [2] To be effective in crop management, these diseases must be identified early and accurately. However, manual inspection by specialists is the foundation of traditional disease detection, which is labor-intensive, time-consuming, and frequently unreliable due to human bias. [3]

Due to their capacity to produce rapid, reliable, and automated diagnoses, computer vision and deep learning techniques for plant disease detection have become increasingly popular in recent years. In image classification tasks, Convolutional Neural Networks (CNNs) have proven to perform better. [3] CNNs are very successful at detecting plant diseases because they directly extract deep features from input images, in contrast to traditional machine-learning techniques that need handcrafted features. [2]

This research paper aims to develop a CNN-based model for detecting tea leaf diseases. The goal is to create an automated system capable of accurately differentiating between healthy and diseased tea leaves based solely on visual characteristics. [2] The study's objective is to provide a practical and applicable solution for real-time tea plantation monitoring, allowing for early disease detection and assisting farmers in reducing losses. [1] This system can assist farmers and agricultural workers in detecting diseases early on, reducing crop damage, and promoting precision agriculture practices. [2]

In this context, our research was guided by two primary questions: (1) How the effectiveness of image processing and AI-based automated systems in accurately identifying tea leaf diseases? (2) To what extent early detection of tea leaf diseases contributes to improving production quality and minimizing losses?

The remainder of this paper is organized as follows. Section II provides a review of related studies on automated tea leaf disease detection and classification. Section III outlines the dataset, preprocessing methods, and model architectures. Section IV presents the experimental setup and results. Section V discusses the findings related to interpretability, evaluates the robustness, and describes the prototype application. Section VI presents a comparison of the proposed framework with existing works. Section VII concludes the study with a summary of the contributions, identified limitations, and suggestions for future research.

## II. RELATED WORKS

To recognize seven tea leaf conditions from photos, Chen et al. [2] created LeafNet, a customized convolutional neural network (CNN). 3,810 leaf photos taken in natural settings formed the dataset, which was later increased to 7,905 photos using augmentation. LeafNet, an improved version of AlexNet with fewer parameters, outperformed SVM 60.6% and MLP 70.8% classifiers based on DSIFT-BOVW features with an average accuracy of 90.2%. CNN's learned features outperformed established techniques in recognition. Further studies could improve consistency and wider applicability. Limitations include moderate accuracy with visually identical symptoms of disease as well as performance decline under various real-world settings.

Yucel and Yıldırım [3] created a tea leaf disease classification system by combining features from three CNN architectures and a linear discriminant classifier. The researchers combined feature maps from multiple well-known CNNs to improve disease recognition and compared their model to seven standard CNNs in the literature. The dataset included tea leaf images labeled by disease class, but the exact size is not specified in the abstract. The results show that the suggested method is more effective than individual CNN models at classifying tea leaf diseases. The study indicates the usefulness of feature fusion but overlooks dataset scale or real-world validation limitations.

With an objective of practical field deployment, Ahammed et al. [4] created a system that used deep learning for identifying deficiencies in nutrition in coffee leaves applying CNN patterns with transfer learning and Grad-CAM visualization. A specific dataset of photos of coffee leaves with six nutritional deficiencies (N, P, K, Ca, Mg, and Fe) was collected, labeled, augmented, and divided for training, validation, and testing. For a web application, the CNN model balanced accuracy and efficiency with the greatest performance 83.32% precision. Decision regions were visualized with Grad-CAM. The next phase focuses on sustainability and multi-crop extension; limitations include a very limited dataset size and variability within real settings.

Kamrul et al. [5] suggested a deep learning-based image classification system utilizing CNN-based algorithms to automate tea leaf categorization for quality rating, replacing traditional texture analysis processes. They examined architectures such as VGG16, sequence models, and Faster R-CNN on a unique tea leaf image dataset, trying to recognize leaf grades or types from images. The models showed high classification accuracies (e.g. 95-96% for VGG16 and Faster R-CNN), indicating how deep CNNs outperform traditional feature methods. Limitations include a small dataset, sensitivity to lighting/background fluctuation, and smaller real-world stability, indicating that future research should extend data and adapt models for field conditions.

Chakraborty et al. [6] created a convolutional neural network (CNN) to automatically classify tea leaf diseases from images, hoping to perform better than traditional classifiers such as SVM and K-NN. They preprocessed and augmented a publicly available tea leaf image dataset with multiple disease and healthy classes before training the CNN to recognize discriminative leaf features. The model achieved 92.6% classification accuracy, outperforming classical methods and indicating CNN's effectiveness for plant disease recognition. Limitations include reliance on curated image datasets and potential performance issues in a wide range of real-world scenarios; future research should expand datasets and investigate advanced architectures for robustness and generalization.

Despite these advances, three challenges remain, namely limited interpretability, insufficient robustness testing, and weak clinical integration [7]. This study addressed these gaps by benchmarking CNN architectures, incorporating interpretability through Grad-CAM and occlusion sensitivity, assessing robustness through adversarial perturbations, and developing a prototype web application.

III. METHODOLOGY

Fig. 1 below illustrates the complete workflow of the proposed classification model. This study represents a deep learning-based approach for detecting Tea leaf diseases using a publicly available dataset. Images were preprocessed through resizing, normalization, and data augmentation techniques that include flipping, rotation, and zooming, to enhance model robustness [8]. We split the dataset into training, testing, and validation sets, after that we applied the oversampling technique on training data to balance the classes [9].

Then we used three pretrained models, Densenet201, MobileNetV2 and InceptionV3. We trained multiple models with various hyperparameters and selected the best model based on the validation metrics. Adversarial training was also performed to improve resistance to noise and perturbations [10]. Finally, we generate visual explanations by using Grad-CAM and highlight the image regions that influence the model's predictions and enhance the interpretability.

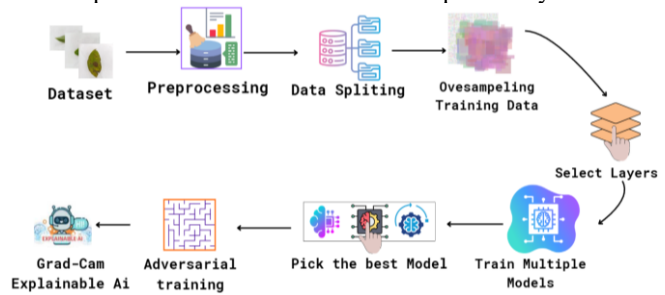

Fig. 1. Methodological framework

A. Dataset

In this research, we used a teaLeafBD dataset shown in Fig. 2.

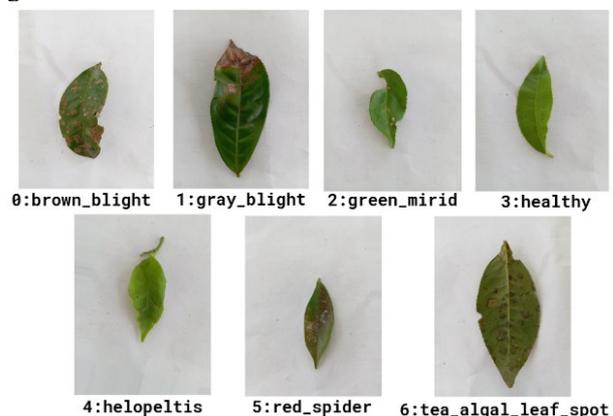

Fig. 2. Sample images from dataset

It contains 5278 RGB high-resolution images of tea leaves, which are categorized into seven classes such as Brown Bligh disease, Gray Blight disease, Green mirid bug disease, Helopeltis disease, Red spider disease, Tea algal leaf spot disease, and Healthy leaf [11]. The images were collected under balanced lighting conditions, varying angles, and backgrounds. All the images were resized to 224x224 pixels for training models, and normalized the pixel values to enhance training stability. The dataset was also divided into three parts of training, testing, and validation. This dataset has set a benchmark for developing accurate and generalized models for Tea Leaf disease classification.

### B. Preprocessing

All images were resized to 224×224 pixels to align with the input specifications of the pre-trained models, and were normalized to the range of [0,1] to facilitate stable optimization. The dataset was divided into subsets comprising 70% training, 20% validation, and 10% testing, ensuring that the evaluation was performed on previously unseen samples. This preprocessing step ensured evenness across all architectures and minimized the impact of variations in the image resolution and contrast.

### C. Model Architectures

To assess model performance across different feature extraction strategies, we benchmarked three well-established CNN architectures.

- *MobileNetV2* advances lightweight CNN design by using inverted residual blocks with depth wise separable convolutions to achieve high accuracy with minimal computational cost [12].
- *InceptionV3* builds on the inception family by factorizing convolutions and using multi-branch modules to improve efficiency and capture multi-scale features [13].
- *DenseNet201* extends the CNN family by introducing densely connected layers, where each layer directly receives inputs from all previous layers to enhance feature reuse and gradient flow [14].

These architectures were selected for their proven success in leaf disease imaging tasks and complementary design characteristics.

### D. Training Setup

Almost the same training method was used for all three models. The Adam optimizer with 0.0001 learning rate was used for DenseNet201 and MobileNetV2 model , and 0.00001 learning rate was used for InceptionV3 model.

TABLE I.  HYPERPARAMETER SETTINGS FOR THE EVALUATED CNN ARCHITECTURES

| Parameter | MobileNetV2 | InceptionV3 | DenseNet201 |
|---|---|---|---|
| Batch size | 32 | 32 | 32 |
| Loss function | Categorical cross-entropy | Categorical cross-entropy | Categorical cross-entropy |
| Learning rate | 0.0001 | 0.00001 | 0.0001 |
| Optimizer | Adam | Adam | Adam |
| Number of epochs | 50 | 50 | 50 |
| Early stopping (patience) | 5 | 10 | 10 |

Categorical cross-entropy is a loss function suitable for tasks with multiple classes for all the models. The batch size was set as 32. For DenseNet201 and InceptionV3, early stopping was used with patience of ten epochs, to prevent overfitting. For MobileNetV2, early stopping was used with a patience of five epochs. The number of epochs was the same for all the models, as shown in Table I.

## IV. RESULTS

This section evaluates the performances of the three CNN models used in this study. In addition, it provides an analysis of the strength of the most effective model.

### A. MobileNetV2

The Fig. 3 the Training and Validation performance graphs of MobileNetV2.

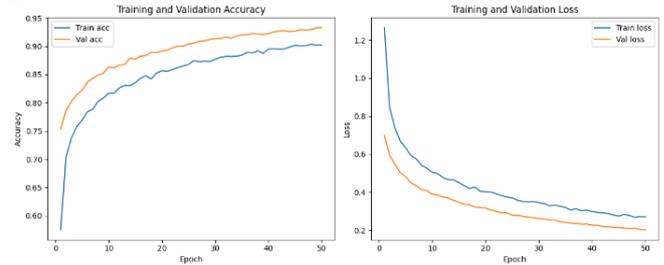

Fig. 3.  Loss for training and validation of MobileNetV2 architecture

The training and validation curves show a steady and consistent learning behaviour. Training accuracy gradually increases from around 58% to 90%, while validation accuracy rises smoothly and even surpasses the training curves, reaching about 93%, which suggests strong generalization due to effective regularization and data augmentation.

The confusion matrix depicted in Fig. 4 indicates a balanced recognition across all classes. Although some misclassifications were observed among the diseases. But these small errors are anticipated because the visual differences between certain tea leaf disease categories are very subtle, especially when the images are captured with a regular camera, where lighting, angle, and texture variations can make adjacent disease types appear very similar.

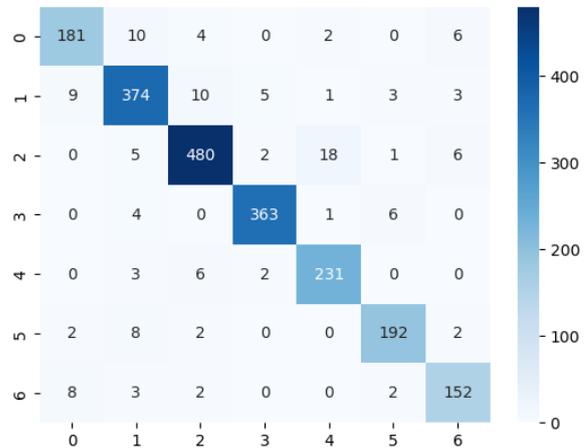

Fig. 4.  Confusion Matrix of MobileNetV2 architecture

### B. InceptionV3

In the scenario Fig. 5 demonstrates effective learning and strong generalization performance through the training and validation of the InceptionV3 architecture. The validation accuracy rises from about 62% to around 92%, closely following the training accuracy, which increases from roughly 30% to about 85% over 50 epochs. In the loss graph, both training and validation loss decrease sharply at the beginning and then continue to decline smoothly over the remaining epochs.

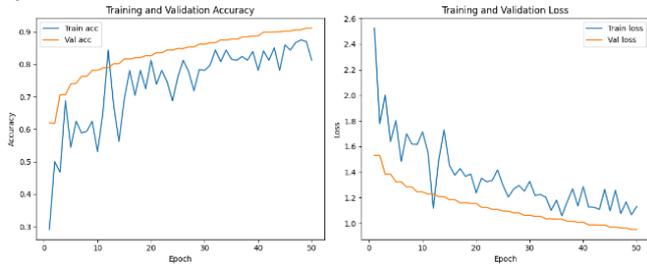

Fig. 5. Training and validation accuracy and loss curve of InceptionV3

The Fig. 6 represents the confusion matrix for InceptionV3, where we can observe a well classification among all the diseases with some expected errors. In comparison to MobileNetV2, InceptionV3 exhibited marginally lower performance in classifying doubtful cases, although most of the classes are balance recognized but some misclassification can be observed in Green mirid bug. Helopeltis and Red spider.

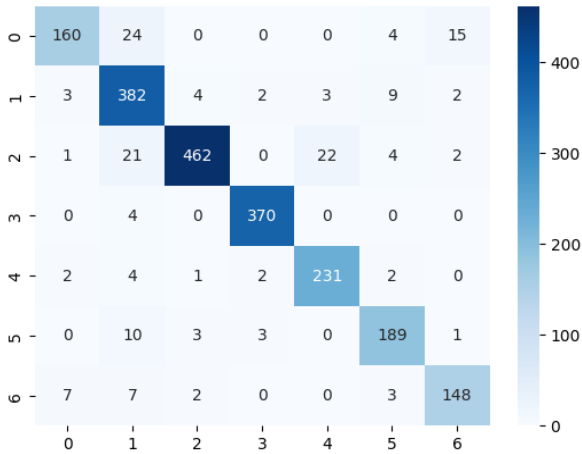

Fig. 6. Confusion Matrix of InceptionV3

### C. DenseNet201

In Fig. 7, the Training and Validation performance graphs of DenseNetV3, illustrates strong and stable learning performance. In the accuracy graph, both training and validation accuracy rise sharply in the first few epochs and then level off around 96% to 100%, indicating fast convergence and excellent generalization with only a small, healthy gap between the two curves.

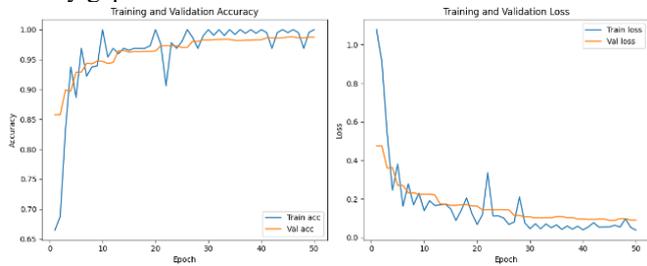

Fig. 7. Training and validation accuracy and loss curve of DenseNet201

In the loss graph, both losses drop quickly at the start and continue decreasing smoothly, with training loss reaching very low values and validation loss remaining close behind. This indicates that the model generalizes effectively to unseen data without showing any significant signs of overfitting and exhibiting the best overall performance among the evaluated models. The confusion matrix presented in Fig. 8 shows high accuracy across most classes, although some misclassifications were observed among the diseases. The superior accuracy and balanced classification outcomes of DenseNet201 establish it as the most reliable model for tea leaf disease classification.

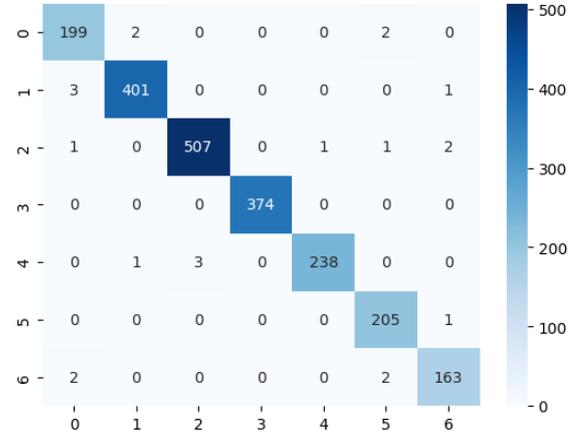

Fig. 8. Confusion Matrix of DenseNet201

### D. Adversarial training

To further assess reliability, adversarial perturbations were applied to DenseNet201, and the results are summarized in Table II. The model consistently demonstrated high validation accuracy across all ε values, achieving a peak accuracy of 98.15% at ε= 0.1. Even at elevated perturbation levels, the performance remained above 97.8%, with only minor variations in loss. These findings confirm that DenseNet201 is not only the most accurate, but also resilient to input noise, thereby reinforcing its suitability for agricultural decision support [15].

TABLE II. ADVERSARIAL ROBUSTNESS EVALUATION OF DENSENET201 UNDER VARYING PERTURBATION LEVELS ($\varepsilon$)

| Epsilon Value ($\varepsilon$) | Validation Loss | Validation Accuracy | Optimal Epochs |
|---|---|---|---|
| 0 | 0.0841 | 0.9896 | 41 |
| 0.1 | 0.0895 | 0.9896 | 50 |
| 0.12 | 0.0757 | 0.9915 | 21 |
| 0.14 | 0.0814 | 0.9910 | 21 |
| 0.16 | 0.0758 | 0.9912 | 23 |
| 0.18 | 0.0750 | 0.9903 | 49 |
| 0.2 | 0.0907 | 0.9872 | 35 |

## V. EXPLAINABLE AI

In addition to being accurate and reliable, models must be easy to understand for use in disease detection applications. To render the deep learning models less like a "black box" we used a method called Grad-CAM [16]. This helps us determine which parts of the leaf affect the model's predictions the most.

### A. Grad-CAM Visualizations

Fig. 9 shows examples of Grad-CAM overlays for different grades. The heat maps focused on the pathological regions of the tea leaves, including lesion centers, boundaries, texture disruptions, and color-altered areas. The model showed minimal attention to healthy or background regions, confirming that predictions were driven by biologically meaningful features rather than noise.

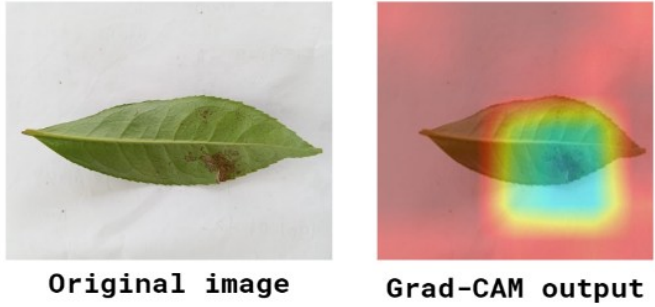

Fig. 9. Tea Leaf Disease identification using Grad-CAM

### B. Occlusion Sensitivity

To demonstrate that our model is easy to understand, we used occlusion sensitivity analysis shown in Fig. 10. Occlusion Sensitivity analysis revealed that the model consistently relied on disease-specific lesion regions [15]. Blocking infected areas—such as necrotic spots, fungal patches, insect feeding marks, and discoloration—caused a significant drop in prediction confidence, confirming that the model's decision-making aligned with the visual pathology of each tea leaf disease.

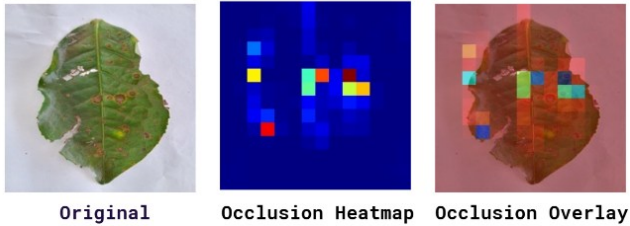

Fig. 10. Tea Leaf Disease identification using Occlusion Sensitivity

Grad-CAM and occlusion sensitivity demonstrated that DenseNet201's predictions were both accurate and interpretable for agricultural experts and farmers. These explainability methods highlighted the exact diseased regions on tea leaves, enabling farmers and field technicians to easily understand why the model made a particular prediction [17]. This transparency increases trust in the system and supports its practical use for real-world tea leaf disease diagnosis.

## VI. Prototype Web Application

To illustrate the practical applicability of the proposed framework, we developed a lightweight web-based application called TeaLeafVision shown in Fig. 11. This interface was designed as a proof-of-concept to demonstrate the integration of automated tea leaf disease detection into the agriculture fields. As shown in Fig. 11, the prototype features a straightforward drag-and-drop interface that allows users to upload tea leaf's images. Upon submission, the system processes the image using the trained DenseNet201 model and returns.

- Predicted tea leaf disease. An associated confidence score quantifies the prediction certainty.
- A Grad-CAM heatmap overlay highlights the pathological regions [15] of the tea leaves that are most influential in the decision of the model.

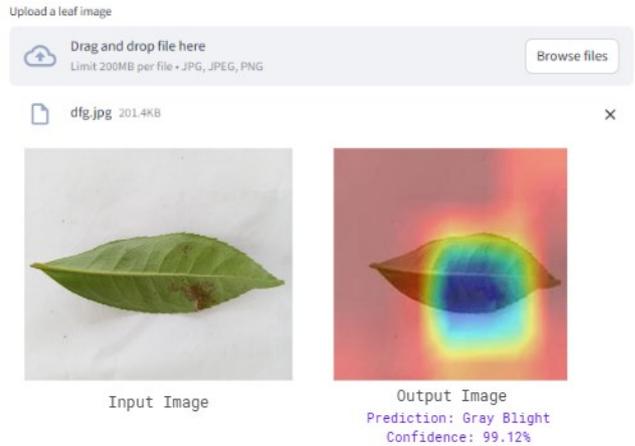

Fig. 11. TeaLeafVision web application for Tea Leaf Disease classification

## VII. Comparison and Discussion

To compare the performance of the proposed TeaLeafVision framework with the existing body of research, we conducted a comparative analysis with the three CNN architectures based on this study as well as with previously documented methodologies for tea leaf disease classification.

### A. Evaluation of CNN Architectures

Table III illustrates a comparative analysis of performance metrics for DenseNet201, MobileNetV2 and inceptionV3.

TABLE III. PERFORMANCE COMPARISON OF EVALUATED CNN ARCHITECTURES

| Model Name | Classes | Precision | Recall | F1-Score |
|---|---|---|---|---|
| DenseNet201 | Brown Blight | 0.97 | 0.98 | 0.98 |
| | Gray Blight | 0.99 | 0.99 | 0.99 |
| | Green mirid bug | 0.99 | 0.99 | 0.99 |
| | Healthy leaf | 1.00 | 1.00 | 1.00 |
| | Helopeltis | 1.00 | 0.98 | 0.99 |
| | Red spider | 0.98 | 1.00 | 0.99 |
| | Tea algal leaf spot | 0.98 | 0.98 | 0.98 |
| MobileNetV2 | Brown Blight | 0.91 | 0.89 | 0.90 |
| | Gray Blight | 0.92 | 0.92 | 0.92 |
| | Green mirid bug | 0.95 | 0.94 | 0.94 |
| | Healthy leaf | 0.98 | 0.97 | 0.97 |
| | Helopeltis | 0.91 | 0.95 | 0.93 |
| | Red spider | 0.94 | 0.93 | 0.94 |
| | Tea algal leaf spot | 0.90 | 0.91 | 0.90 |
| InceptionV3 | Brown Blight | 0.92 | 0.79 | 0.85 |
| | Gray Blight | 0.85 | 0.94 | 0.89 |
| | Green mirid bug | 0.98 | 0.90 | 0.94 |
| | Healthy leaf | 0.98 | 0.99 | 0.99 |
| | Helopeltis | 0.90 | 0.95 | 0.93 |
| | Red spider | 0.90 | 0.92 | 0.91 |
| | Brown Blight | 0.92 | 0.79 | 0.85 |

DenseNet201 exhibited the highest test accuracy 99%, which is significantly higher than MobileNetV2 and InceptionV3 having accuracy of 94% and 92% respectively. The performance of these models were consistently observed across precision, recall, and F1-score metrics.

*B. Benchmarking Against Prior Work*

We conducted a comparative analysis of TeaLeafVision and existing studies on tea leaf disease classification. As Illustrated in Table IV, the DenseNet201 model achieved an accuracy of 99%, surpassing the several previous methodologies, For instance, Soeb et al. [1] reported an accuracy of 97.3% using YOLOv7 on the tea leaf dataset, whereas Chen et al. [2] achieved an accuracy of 90.16% using LeafNef. Ahammed et al. [4] achieved 83.32% accuracy using the Xception model. Compared to these models TeaLeafVision consistently demonstrated superior accuracy and minimized class-level misclassification. The significant distinction in the existing work is that numerous studies have particularly concentrated on accuracy, often neglecting the aspect of interpretability and robustness. But TeaLeafVision has illustrated high accuracy while effectively ensuring transparency and stability, which is crucial for field implementation.

TABLE IV. COMPARISON OF MEDIVISION WITH PREVIOUS WORKS ON TEA LEAF DISEASE CLASSIFICATION

| Author | Method | Accuracy (%) |
|---|---|---|
| Soeb et al. [1] | YOLOv7 | 97.3 |
| Chen et al. [2] | LeafNet | 90.16 |
| N. Yucel et al. [3] | ShuffleNet | 91.30 |
| Ahammed et al. [4] | Xception | 83.32 |
| Karmokar et al. [9] | Neural Network Ensemble | 91 |
| Proposed Model | Inception3 | 92 |
| | MobileNetV2 | 94 |
| | DenseNet201 | 99 |

*C. Discussion of Limitations and Future Work*

This comparative study provides three principle contributions. DenseNet201 exhibited the most reliable classification performance among the evaluated architectures. Second, TeaLeafVision exceeds several documented approaches in terms of accuracy while also combining interpretability and robustness. Third, Grad-CAM analysis verified that the model depends on the effects spot and damages on the leaf. Among several limitations, the dataset was small in size with class imbalance. Moreover the study relied on a single public dataset without external validation. Although adversarial robustness tests confirmed the noise resistance, further investigation is required to assess the research domain shifts across the imaging centers and devices. Future works should focus on model optimization and data set expansion involving real life field implementation.

## VIII. CONCLUSION

This study introduced TeaLeafVision, a deep learning tool that helps classify the disease of tea leaves. It tested three types of neural networks on a task that involved healthy tea leaf and six tea leaf disease. DenseNet201 performed the best with the test accuracy of 99% and was robust against small changes in the images. Grad-CAM images and occlusion sensitivity tests were used to simplify the interpretation of the results. These results show that the detections of the models are based on the affected areas of the leaf. TeaLeafVision application was created to demonstrate how it can be used as a tool to help farmers and agricultural experts. Early detection can help farmers and experts to take early caution which will improve the tea leaf production quality and minimize the loss. Here we get the answer to our second research question, early detection of tea leaf diseases can significantly improve production quality and minimizing losses. Although the results of this study are promising, several limitations remain. Despite all the challenges, TeaLeafVision effectively uses AI to access tea leaf disease ensuring high accuracy as well as clarity and robustness.